\documentclass[conference]{IEEEtran}
\IEEEoverridecommandlockouts
\usepackage{cite}
\usepackage{amsmath,amssymb,amsfonts}
\usepackage{algorithmic}
\usepackage{graphicx}
\usepackage{textcomp}
\usepackage{xcolor}
\usepackage{tabularx}
\usepackage{subfigure}

\def\BibTeX{{\rm B\kern-.05em{\sc i\kern-.025em b}\kern-.08em
    T\kern-.1667em\lower.7ex\hbox{E}\kern-.125emX}}
\begin{document}

\title{Infant Care Video Dataset for Classification of Interventions Using Transformers}

%\author{\IEEEauthorblockN{Anonymous Submission}}
\author{\IEEEauthorblockN{Igor Bogdanov}
\IEEEauthorblockA{\textit{Department of Systems and Computer Engineering} \\
\textit{Carleton University}\\
Ottawa, Canada \\
igorbogdanov@cmail.carleton.ca}
\and
\IEEEauthorblockN{James Green}
\IEEEauthorblockA{\textit{Department of Systems and Computer Engineering} \\
\textit{Carleton University}\\
Ottawa, Canada \\
jrgreen@sce.carleton.ca}}

\maketitle

\begin{abstract}
Healthcare documentation in the neonatal intensive care unit (NICU) presents significant challenges, with nurses spending approximately 25\% of their time on record-keeping, while up to 60\% of interventions remain undocumented. Motivated by the need to detect interventions from video automatically, we present the Infant Care Video Dataset (ICVD), a collection of 4,144 videos spanning 12 simulated intervention classes designed for developing automated documentation systems. Our manikin-based approach systematically varies conditions, such as camera angle and clinician skin tone, while ensuring privacy compliance. Using video transformer architectures (TimeSformer and MotionFormer), we establish strong baseline performance (93.97\% and 93.17\% top-1 accuracy) among the 12 infant care classes. Our ablation study comparing temporal models with a framewise approach (23.17\% accuracy) demonstrates a 70.80\% performance gap, validating the need for temporal modeling. The ICVD provides a foundation for developing automated documentation systems to reduce clinical burden in neonatal care environments and improve existing practices.
\end{abstract}


\section{Introduction}
Neonatal intensive care units (NICUs) face a critical challenge balancing patient care with documentation requirements. Documentation consumes 25-50\% of clinicians' time \cite{b1}. Nurses perform an average of 35-45 direct care interventions per infant per 12-hour shift, with up to 60\% of routine interventions going undocumented due to workflow constraints \cite{b2}. Computer vision-based automation offers a promising solution, but development is stalling due to the absence of specialized datasets for this domain \cite{b3}.

Privacy regulations pose exceptional barriers in neonatal healthcare. While medical video datasets exist for surgical procedures, no publicly available datasets capture the routine care procedures specific to NICU environments \cite{b4, b5, b6}. This absence represents a fundamental constraint on developing automated documentation systems.

This paper presents three contributions: (1) the Infant Care Video Dataset (ICVD), comprising 4,144 videos of simulated care spanning 12 standardized NICU procedures; (2) performance benchmarks established through transfer learning with state-of-the-art video transformer architectures; and (3) a methodological framework for intervention classification in privacy-sensitive healthcare domains. Our manikin-based approach preserves essential visual characteristics while addressing ethical and privacy constraints, enabling broader data access than datasets requiring restricted protocols.

The ICVD targets the documented problem of intervention under-reporting in clinical settings, where correct reporting rates for routine procedures can fall as low as 31\% \cite{b2}. We identified the selected intervention classes through consultation with practicing NICU nurses. These classes represent high-frequency activities performed repeatedly throughout shifts but inconsistently documented due to workflow constraints.

Our experimental results demonstrate that temporal information is critical for accurate intervention classification, with a 70.80\% performance gap between frame- and video-based approaches. The TimeSformer \cite{b7} and MotionFormer \cite{b8} models achieved 93.97\% and 93.17\% top-1 accuracy, respectively, validating our dataset design decisions for automated documentation systems.

\section{Related Work}
\subsection{Medical Video Datasets}
In the surgical domain, several high-quality video datasets have established benchmarks for workflow analysis, including JIGSAWS \cite{b9} and Cholec80 \cite{b10}. However, neonatal care reveals a significant gap in procedure-focused datasets. Most existing neonatal collections target physiological monitoring rather than care interventions, such as the NBHR dataset \cite{b4} and AUB Neonatal Vitals dataset \cite{b5}.

Most relevant to this study is the recent NICU Care Activities dataset \cite{b3}, which contains 289 hours of real infant footage (three types of interventions), but is inaccessible due to privacy issues. ICVD addresses these limitations by providing 12 intervention classes with a manikin-based approach that eliminates privacy concerns while prioritizing accessibility.

\subsection{Action Recognition in Healthcare}
Video-based action recognition in healthcare has evolved significantly, with transformer-based architectures increasingly replacing CNNs due to their ability to capture long-range temporal dependencies \cite{b16}. Recent applications include surgical workflow recognition, though neonatal care presents unique challenges \cite{b11}.

Majeedi et al. \cite{b3} applied temporal transformer models to detect three different caregiving activities in NICU videos, while Hajj-Ali et al. \cite{b12} used vision transformers with depth cameras to detect the presence of clinicians in the patient care space. Using a privacy-preserving dataset, our work extends classification to a broader set of intervention classes.

\subsection{NICU Documentation}
NICU documentation faces unique challenges due to continuous care and frequent interventions. In one study that examined documentation of patient repositioning interventions, documentation compliance was shown to be as low as 31\%. Automated systems improved compliance to 82\% \cite{b2}. Underreporting creates a limited picture and makes it hard to evaluate the impacts of caregiving on the outcomes \cite{b13}.

NICU nurses may need to record hundreds of data points per shift, often navigating complex interfaces. In most cases, the nurse must leave the patient care space to complete data entry. Ultimately, documentation can consume hours \cite{b14}, reducing time for direct patient care and potentially impacting clinical decision-making when vital information is missed \cite{b15}.

\section{Dataset Design and Collection}
\subsection{Design Principles}
Five key principles guided the ICVD:
\begin{enumerate}
    \item \textbf{Clinical Relevance:} Intervention classes were selected based on frequency and documentation importance in NICU settings.
    
    \item \textbf{Action Discreteness:} Interventions were designed to be temporally distinct and recognizable through hand movements.
    
    \item \textbf{Hierarchical Complexity:} ICVD includes atomic actions and composite procedures, allowing models to learn relationships between simple and complex interventions.
    
    \item \textbf{Standardization with Variability:} We incorporated controlled variability in lighting, backgrounds, hand coverings, and recording devices.
    
    \item \textbf{Ethical Accessibility:} Using a manikin and showing only the clinician's hands creates videos that can be widely distributed without privacy constraints.
\end{enumerate}

\subsection{Procedure Selection}
We identified 12 distinct intervention classes representing the most frequent hands-on care activities that require documentation but are often underreported:

\begin{itemize}
    \item \textbf{Giving/Taking Pacifier:} Non-nutritive sucking interventions critical for comfort \cite{b19}.
    
    \item \textbf{Diaper-related:} Changing, applying, and removing diapers, frequent daily intervention \cite{b3}.
    
    \item \textbf{Feeding:} Directly impacting growth outcomes and requiring detailed documentation \cite{b3}.
    
    \item \textbf{Crib handling:} Removing and replacing the patient from the crib or overhead warmer bed, marking significant transitions \cite{b3}.
    
    \item \textbf{Hygiene:} Wiping body and wiping face, typically performed multiple times daily \cite{b3}.
    
    \item \textbf{Temperature management:} Covering and uncovering the patient for thermoregulation \cite{b20}.
\end{itemize}

The Dataset includes atomic actions and composite sequences (e.g., diaper change that incorporates removing and replacing diapers).

\subsection{Recording Methodology}
\begin{figure}[t]
\centering
\includegraphics[width=0.75\columnwidth]{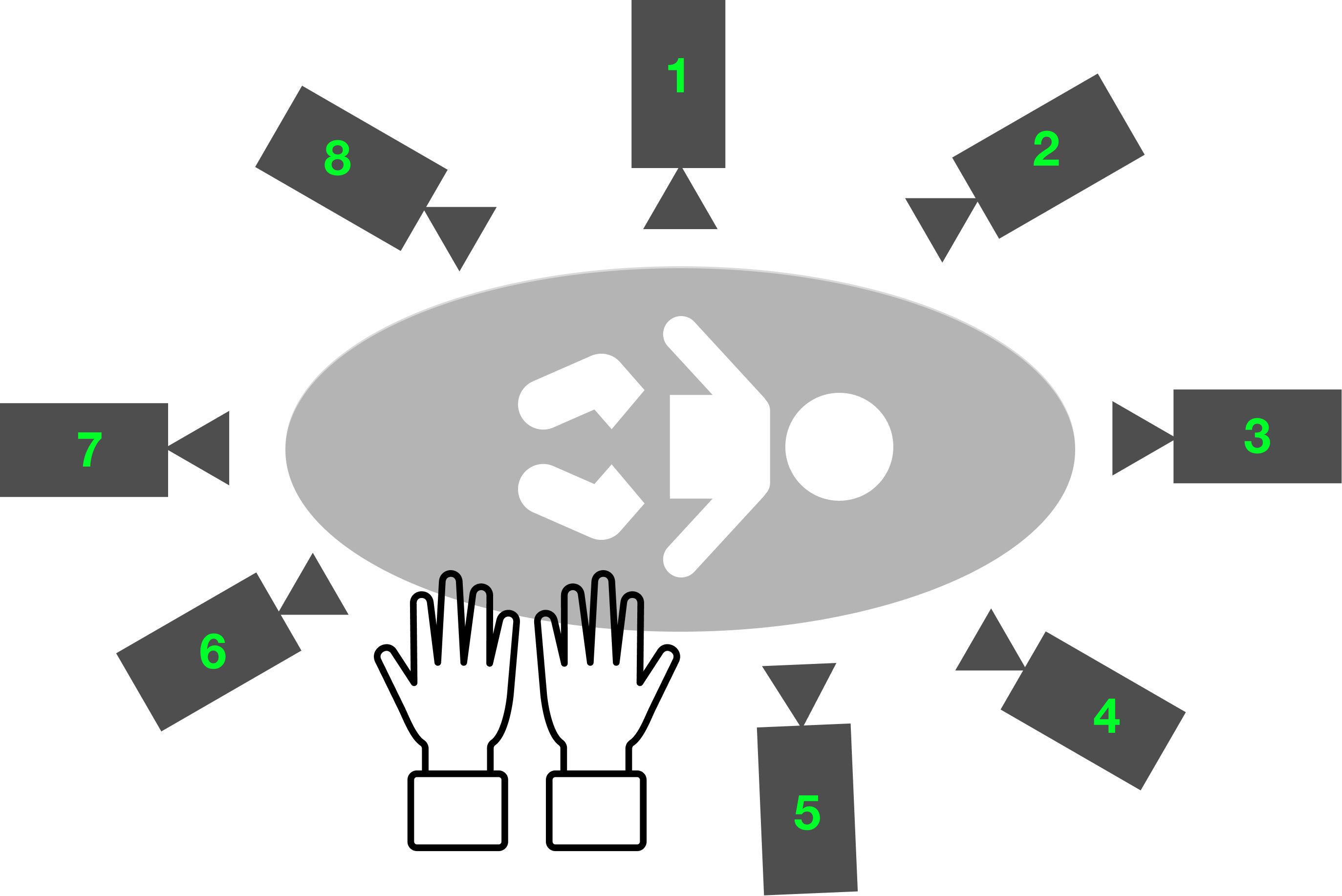}
\caption{Camera configuration used for dataset recording. Eight cameras were positioned around the infant manikin in the crib. The cameras included several smartphone models and baby monitors to capture diverse perspectives and recording quality, representing real-world monitoring conditions.}
\label{fig:camera_setup}
\end{figure}

Our recording configuration used eight cameras arranged around the crib (Fig.~\ref{fig:camera_setup}), mounted at various angles at similar distances from the crib center. The setup included two iPhone 11 Pro Max devices, an iPhone 8 Plus, two Arenti 360° View 5G WiFi Pet monitors, an Arenti 4MP Baby Monitor, a Google Pixel 7 Pro, and a Samsung Galaxy S22 Ultra.

Including diverse cameras in the Dataset enables the development of robust intervention classification systems that function independently of the specific camera model. Furthermore, recording sessions systematically varied other conditions: natural/artificial lighting, light/dark crib sheets, and five hand covering types.

The recording process generated approximately 48 hours of unedited footage. This raw footage was reviewed, segmented, and edited during post-production using Final Cut Pro into 3- to 40-second clips. Each clip captured a single, complete instance of an intervention and was isolated to show only the hands performing the intervention and the manikin, maintaining the privacy-preserving design of the Dataset. 

The post-production phase yielded 4,144 individual video clips from an initial target of 4,160 with a rigorous quality control process that resulted in a 99.6\% acceptance rate. Quality control included assessment of intervention completeness, visual clarity, procedural accuracy, and background consistency, resulting in only 16 videos (0.4\%) being rejected from the initial recordings. 

\section{Dataset Properties and Organization}

\subsection{Technical Specifications}
All videos in the ICVD were standardized to ensure consistent technical specifications while preserving essential visual details for intervention classification. Each recording is provided in .mov or .mp4 format with H.264 video encoding at approximately 270 kbps bitrate. The videos are uniformly presented in landscape orientation and were cropped and resized to standardized dimensions of 456×256 pixels, while maintaining the original aspect ratio.

The frame rate of each clip varies between 24 and 60 fps, depending on the camera output, and provides sufficient temporal resolution to capture the motion dynamics of interventions. Audio tracks were removed during processing to eliminate privacy concerns and reduce file sizes.

The mean duration of clips is approximately 17 seconds, with longer durations corresponding to more complex procedures such as diaper changing. The Dataset occupies approximately 2.21 GB, with individual clips averaging 570-600 KB each.

\subsection{Temporal Characteristics}
\begin{figure}[h!]
\centering
\setlength{\subfigcapskip}{5pt}
\setlength{\subfigtopskip}{0pt}
\setlength{\subfigbottomskip}{0pt}
\setlength{\subfigcapmargin}{1pt}
\subfigure[t=0s]{%
  \includegraphics[width=0.22\columnwidth]{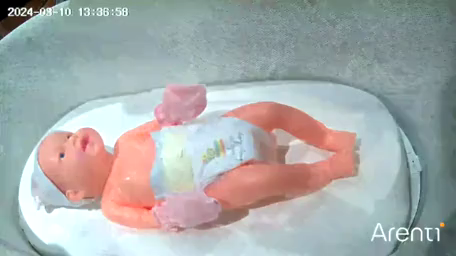}%
}%
\hspace{1pt}%
\subfigure[t=1s]{%
  \includegraphics[width=0.22\columnwidth]{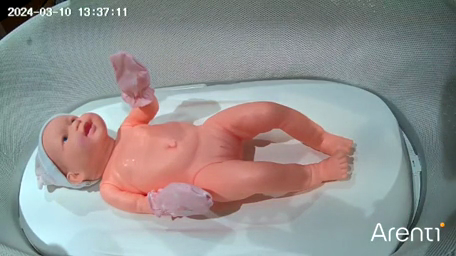}%
}%
\hspace{1pt}%
\subfigure[t=2s]{%
  \includegraphics[width=0.22\columnwidth]{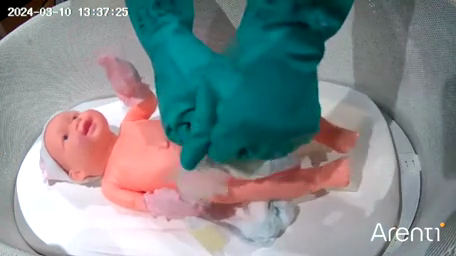}%
}%
\hspace{1pt}%
\subfigure[t=3s]{%
  \includegraphics[width=0.22\columnwidth]{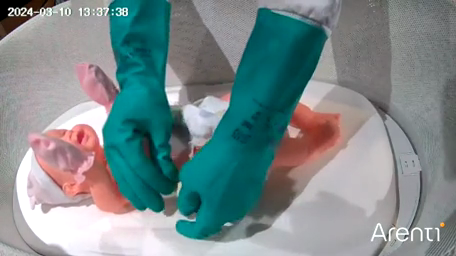}%
}%
\caption{Sample frames from a \textbf{diaper-change} intervention showing: (a) preparing, (b) removing, (c) starting diaper application, and (d) applying a fresh diaper.}
\label{fig:sample_frames}
\end{figure}

The temporal dimension is critical for differentiating interventions that appear visually similar in static frames. For example, \textbf{covering} and \textbf{uncovering} involve identical visual elements but in reverse temporal order; static image analysis would be insufficient to distinguish between these classes. Similarly, several interventions contain overlapping motion patterns requiring temporal context for proper classification, a challenge noted in recent healthcare action recognition research \cite{b16}. Fig.~\ref{fig:sample_frames} provides samples from \textbf{diaper-change} intervention that involves two similarly looking atomic actions: \textbf{removal} and \textbf{replacement of the diaper}.

To validate the importance of temporal information, we conducted an ablation study comparing frame-based models with the whole video transformer approach, confirming that temporal modeling substantially improved classification accuracy, particularly for visually similar intervention pairs. This validation reinforces our design decision to emphasize complete temporal sequences rather than representative keyframes, consistent with findings in surgical workflow analysis \cite{b16, b7}.

\subsection{Data Variability}
To ensure robustness, we incorporated controlled variability (Fig.~\ref{fig:variability_factors}):

\begin{figure}[t]
\centering
\includegraphics[width=\columnwidth]{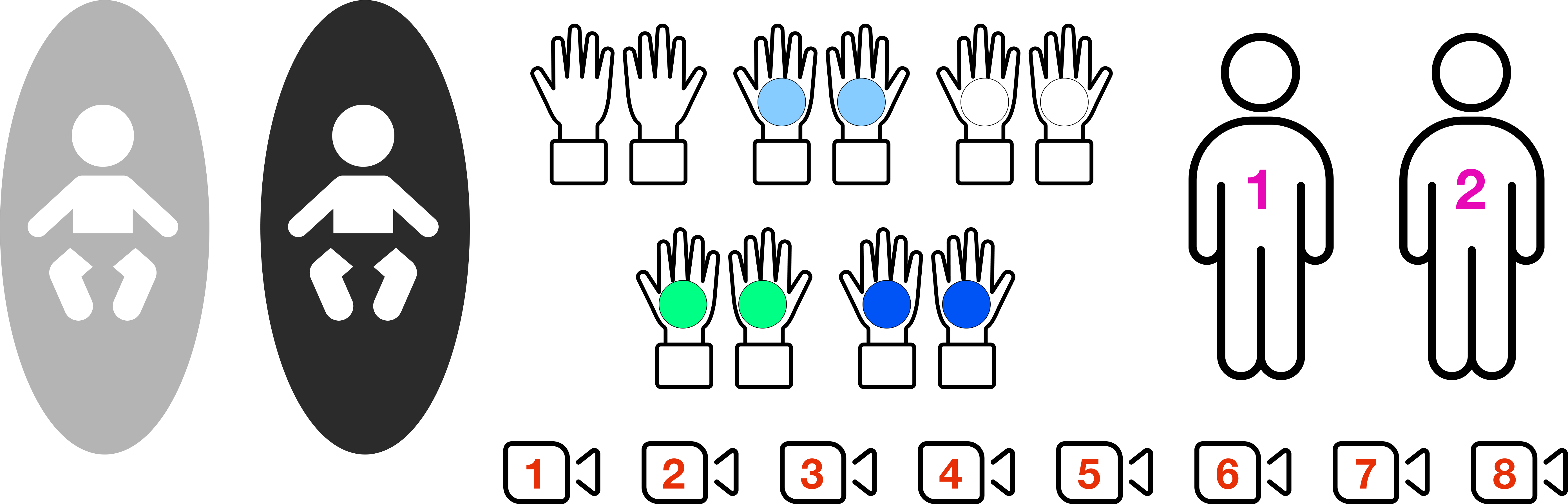}
\caption{Systematic variability factors in the ICVD dataset. Left: Two background conditions. Center: Five hand covering types. Right: Two performers and eight camera perspectives.}
\label{fig:variability_factors}
\end{figure}

\begin{enumerate}
    \item \textbf{Backgrounds:} Light/dark crib sheets.
    
    \item \textbf{Hand coverings:} Five different colours of gloves reflecting clinical practice.
    
    \item \textbf{Performers:} One male and one female with varying hand sizes and techniques.
    
    \item \textbf{Camera perspectives:} Eight positions providing varied viewing angles.
    
    \item \textbf{Lighting:} Both artificial and natural lighting scenarios.
\end{enumerate}

\begin{figure}[t]
\centering
\includegraphics[width=\columnwidth]{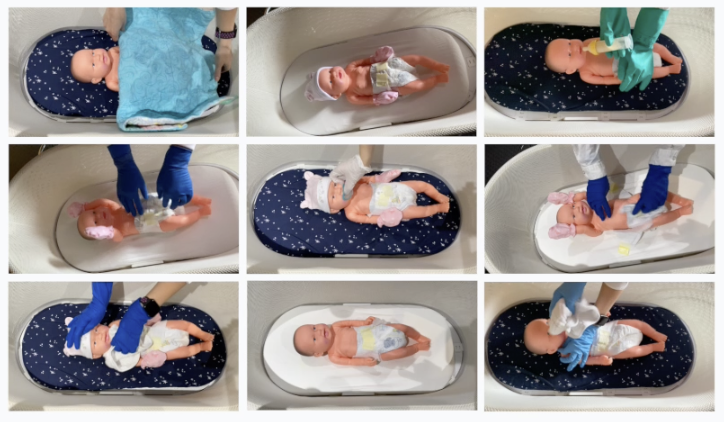}
\caption{Sample frames showcase the visual diversity across intervention types, camera angles, hand coverings, and background conditions. Top row (left to right): \textbf{blanket-cover} with bare hands, \textbf{putting baby to crib}, feeding with green gloves. Middle row: \textbf{diaper-remove} with blue gloves on light sheet, \textbf{pacifier-give} with white gloves on dark sheet, \textbf{diaper-change} with dark blue gloves. Bottom row: \textbf{wiping-face} with blue gloves, \textbf{blanket-uncover} on white sheet, \textbf{wiping-face} with light blue gloves on dark sheet. Note the systematic variation in hand coverings (bare hands, white, green, dark blue, light blue gloves) and background conditions (dark and light sheets).}
\label{fig:intervention_samples}
\end{figure}

Fig. \ref{fig:intervention_samples} shows sample frames from different intervention classes in the ICVD dataset.

\subsection{Organization and Distribution}
The Dataset follows a three-level hierarchical organization: split level (70-15-15 train-validation-test ratio, reflected in Table \ref{tab:table_split}), class level (12 intervention categories), and lighting condition level (dark/light backgrounds). The ICVD maintains class balance with most classes containing 310-340 recordings (Table~\ref{tab:class_distribution}). The \textbf{wiping-body} class contains more videos (610) since it was repeated both in the presence and absence of a diaper to capture a greater variety of techniques.

\begin{table}[h!]
  \centering
  \caption{Dataset split for model training}
  \small
  \begin{tabular}{|l|c|}
    \hline
    \textbf{Split} & \textbf{Number of videos} \\
    \hline
    Total & 4144 \\
    \hline
    Train & 2892 \\
    \hline
    Validation & 624 \\
    \hline
    Test & 631 \\
    \hline
    \end{tabular}   
    \label{tab:table_split}
\end{table}

\begin{table}[h!]
  \centering
  \caption{Number of videos by class and background type}
  \small
  \begin{tabular}{|l|c|c|c|c|}
    \hline
    \textbf{Classes} & \textbf{Index} & \textbf{Light} & \textbf{Dark} & \textbf{Total} \\
    \hline
    Giving Pacifier & 7 & 172 & 162 & 334 \\
    \hline
    Changing Diaper & 10 & 160 & 145 & 305 \\
    \hline
    Taking Pacifier & 5 & 163 & 158 & 321 \\
    \hline
    Feeding & 6 & 163 & 152 & 315 \\
    \hline
    Removing from crib & 1 & 165 & 172 & 337 \\
    \hline
    Replacing infant in crib & 8 & 166 & 173 & 339 \\
    \hline
    Wiping body, w/wo diaper & 0 & 313 & 297 & 610 \\
    \hline
    Wiping face & 4 & 165 & 159 & 324 \\
    \hline
    Covering & 2 & 162 & 157 & 319 \\
    \hline
    Uncovering & 9 & 158 & 157 & 315 \\
    \hline
    Applying Diaper& 3 & 156 & 154 & 310 \\
    \hline
    Removing Diaper & 11 & 160 & 155 & 315 \\
    \hline
  \end{tabular}
  \label{tab:class_distribution}
\end{table}

\subsection{Access Protocol and Reproducibility}
We established a controlled access protocol for the ICVD to promote reproducible research while ensuring ethical use. Researchers can request access via https://www.infantcaredataset.org. Approved users are granted a non-exclusive license to maximize ICVD's impact and facilitate innovation in healthcare documentation automation. 
%This approach balances accessibility with appropriate oversight, similar to protocols used for other healthcare datasets \cite{b18}.

\section{Baseline Experiments}
\subsection{Methods}
Two video transformer architectures, pretrained on Kinetics-400 \cite{b17}, were fine-tuned on the training dataset to establish state-of-the-art baseline performance. Our experiments focused on two distinct attention mechanisms: TimeSformer's (TSF) divided space-time attention \cite{b7} and MotionFormer's (MF) trajectory attention \cite{b8}, both representing different approaches to modeling spatiotemporal relationships in video data \cite{b16}. The following fine-tuning pipeline was used for both methods:

\begin{enumerate}
    \item \textbf{Preprocessing:} Input videos were processed using uniform temporal sampling. TSF extracted 20 frames per clip (stride of 32), while MF used 16 frames per clip (stride of 4), with the same target framerate of 30 FPS. Frames were resized to 224×224 pixels using center cropping during evaluation and random cropping during training (scale jittering range 0.8-1.0). Inputs were normalized with mean=[0.45, 0.45, 0.45] and standard deviation=[0.225, 0.225, 0.225] for TSF, while MF used mean=[0.5, 0.5, 0.5] and standard deviation=[0.5, 0.5, 0.5]. Additional augmentations included random horizontal flipping (p = 0.5) and color jittering for MF, which was disabled for TSF.
    
    \item \textbf{Model architecture:} TSF utilizes a ViT-base configuration with 12 transformer layers, 12 attention heads, embedding dimension of 768, and patch size of 16×16 pixels \cite{b7}. The divided space-time attention mechanism first computes self-attention along the spatial dimension for each frame, followed by temporal attention across frames \cite{b7}. MF employs 12 transformer blocks with trajectory attention that explicitly models motion tubes by tracking corresponding tokens across frames, with an embedding dimension of 768, 12 attention heads, and a patch size of 16×16 pixels \cite{b8}. Both architectures have proven effective for long-range temporal modeling in healthcare videos \cite{b16}.
    
    \item \textbf{Optimization:} TSF was fine-tuned using the SGD optimizer with a base learning rate of 0.005, momentum of 0.9, and weight decay of 0.0001, following the optimization strategy in \cite{b7}. MF used the AdamW optimizer with a lower base learning rate of 0.0001 and weight decay of 0.05, reflecting its different pretraining approach and architecture characteristics \cite{b8}. Both models used a step-based learning rate schedule with relative learning rates at predefined epochs. TSF was trained for 15 epochs (approximately 3 hours of computation), while MF required 35 epochs (approximately 2 hours) due to its lower learning rate and different attention mechanism. All models were trained on a single NVIDIA A100 40GB GPU with a batch size of 8.
    
    \item \textbf{Ablation design:} To assess the importance of temporal information in intervention classification, we designed an ablation study comparing the full TSF model against a framewise (FW) version that processed video frames using space-only attention and single-frame input. \cite{b16}. 
\end{enumerate}

We computed class-balanced macro-averaged metrics for evaluation, including top-1/top-5 accuracy, precision, recall, and F1-scores.

\subsection{Results}
Both video transformer models achieved exceptional performance (Table \ref{tab:model_performance}). TSF reached 93.97\% top-1 and 100\% top-5 accuracy on the test set, with similar results from MF (93.17\% top-1; 99.84\% top-5).  
\begin{table}[h!]
  \centering
  \caption{Model Performance Metrics On The Test Set}
  \scriptsize
  \begin{tabular}{|l|c|c|c|c|c|}
    \hline
    \textbf{Model} & \textbf{Top-1 (\%)} & \textbf{Top-5 (\%)} & \textbf{Precision (\%)} & \textbf{Recall (\%)} & \textbf{F1 (\%)} \\
    \hline
    TSF & 93.97 & 100.00 & 94.11 & 93.29 & 92.92 \\
    \hline
    MF & 93.17 & 99.84 & 93.73 & 92.32 & 91.29 \\
    \hline
    FW & 23.17 & 65.87 & 23.08 & 23.00 & 19.45 \\
    \hline
  \end{tabular}
  \label{tab:model_performance}
\end{table}

Our ablation study revealed that the framewise (FW) approach achieved only 23.17\% top-1 accuracy, demonstrating a 70.80\% performance gap compared to the temporal model. This gap provides compelling evidence for the critical importance of temporal information in intervention classification.

\subsubsection{Class-Level Performance Analysis}
\begin{figure}[h!]
\centering
\includegraphics[width=\columnwidth]{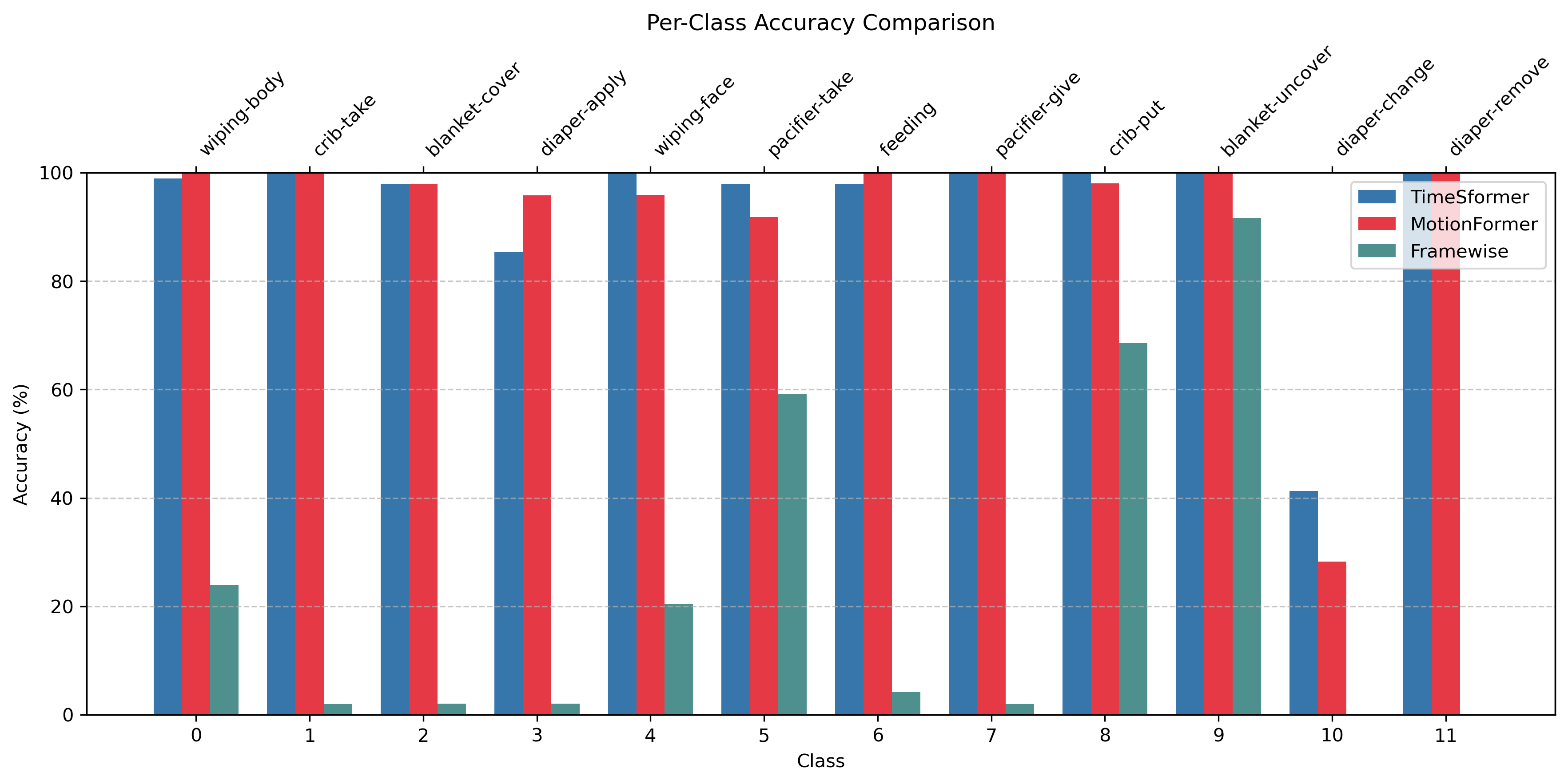}
\caption{Per-class accuracy comparison between models. Temporal models achieve near-perfect accuracy on most classes, unlike the framewise model.}
\label{fig:per_class_comparison}
\end{figure}

Examining the performance at the class level highlights specific patterns in model capabilities and limitations. The per-class analysis of TSF and MF results reveals outstanding model performances for most interventions. For example, for TSF, as shown in Fig.~\ref{fig:per_class_comparison}, six classes achieved perfect 100\% accuracy: \textbf{crib-take} (removing from crib), \textbf{wiping-face}, \textbf{pacifier-give}, \textbf{crib-put} (replacing infant in the crib), \textbf{blanket-uncover}, and \textbf{diaper-remove}. Three additional classes achieved excellent performance above 97\%: \textbf{wiping-body} (98.91\%), \textbf{blanket-cover} (97.96\%), and \textbf{pacifier-take} (97.96\%). The \textbf{feeding} class performed at 97.92\% accuracy.

\subsubsection{Temporal Context and Hierarchical Activities}
\begin{figure}[h!]
\centering
\includegraphics[width=\columnwidth]{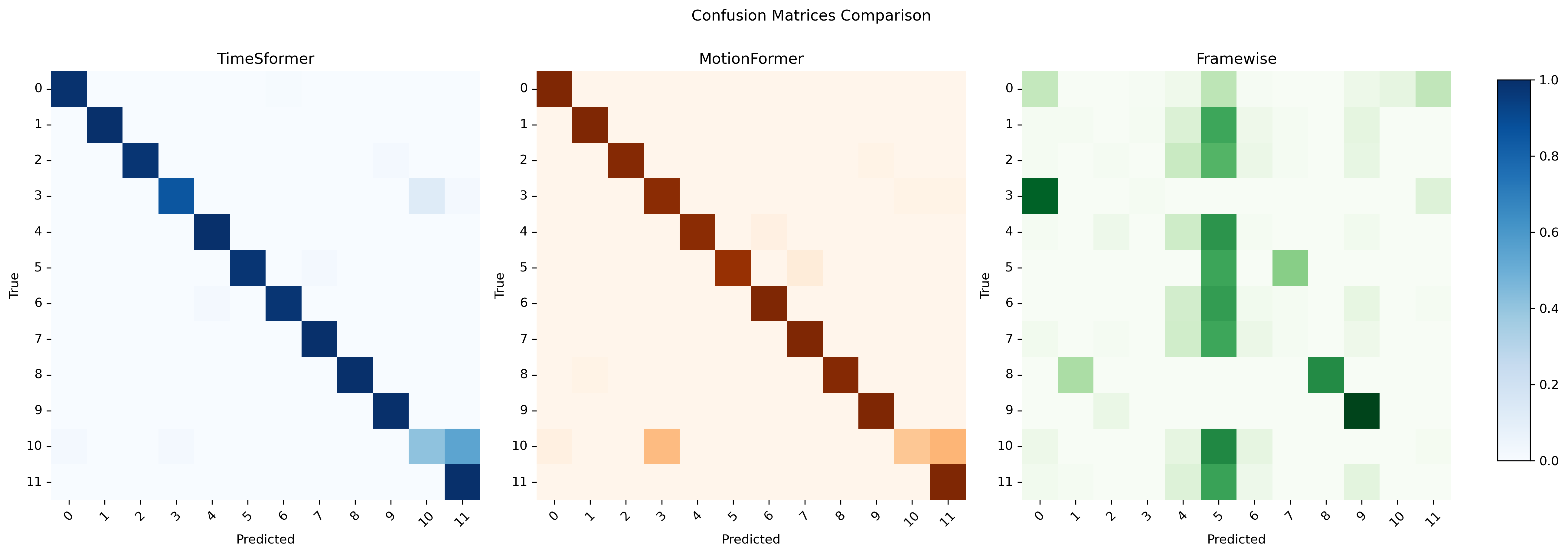}
\caption{Normalized confusion matrices comparing models. Left: TimeSformer shows near-perfect classification for most classes. Middle: MotionFormer exhibits similar patterns. Right: The framewise model shows substantial confusion across classes.}
\label{fig:confusion_comparison}
\end{figure}

Two classes exhibited notable challenges for the model. The \textbf{diaper-apply} class achieved 85.42\% accuracy, with most confusions occurring with \textbf{diaper-change} (6 instances) and \textbf{diaper-remove} (1 instance). Most significant drop in performance can be observed for the \textbf{diaper-change} class at only 41.30\% accuracy, with 25 cases being confused with \textbf{diaper-remove}, as visualized in the confusion matrix (Fig.~\ref{fig:confusion_comparison}, left). This pattern of confusion is logically consistent, as \textbf{diaper-change} includes elements of both \textbf{removing and applying a diaper}, making it particularly challenging for the model to identify \textbf{diaper-change} as a distinct procedure rather than as its component actions. An analogous phenomenon has also been observed in composite surgical activities \cite{b16}.

MF shows a similar pattern (Fig.~\ref{fig:confusion_comparison}, middle and Fig.~\ref{fig:per_class_comparison}), with \textbf{diaper-change} achieving only 28.26\% accuracy. This consistently poor performance across both architectures on this particular class highlights the challenge of distinguishing composite activities from their parts, a known difficulty in hierarchical action recognition tasks \cite{b11, b3}.

The framewise model particularly struggled with temporally-defined class pairs: \textbf{covering/uncovering} and \textbf{diaper-apply/diaper-remove} were frequently confused as these classes involve identical visual elements performed in reverse temporal order. Unlike the complete temporal model, the framewise approach could not disambiguate these visually similar actions, resulting in near-random classification between such pairs. The framewise model's low precision (23.08\%), recall (23.00\%), and F1 score (19.45\%) further confirm its inability to handle the classification task without temporal context. Classes with more distinctive spatial features (like \textbf{crib-put} at 68.63\% and \textbf{blanket-uncover} at 91.67\%) achieved somewhat better performance but still fell far short of the temporal model's accuracy, demonstrating the limits of static frame analysis for dynamic activities.

\begin{figure}[h!]
\centering
\includegraphics[width=\columnwidth]{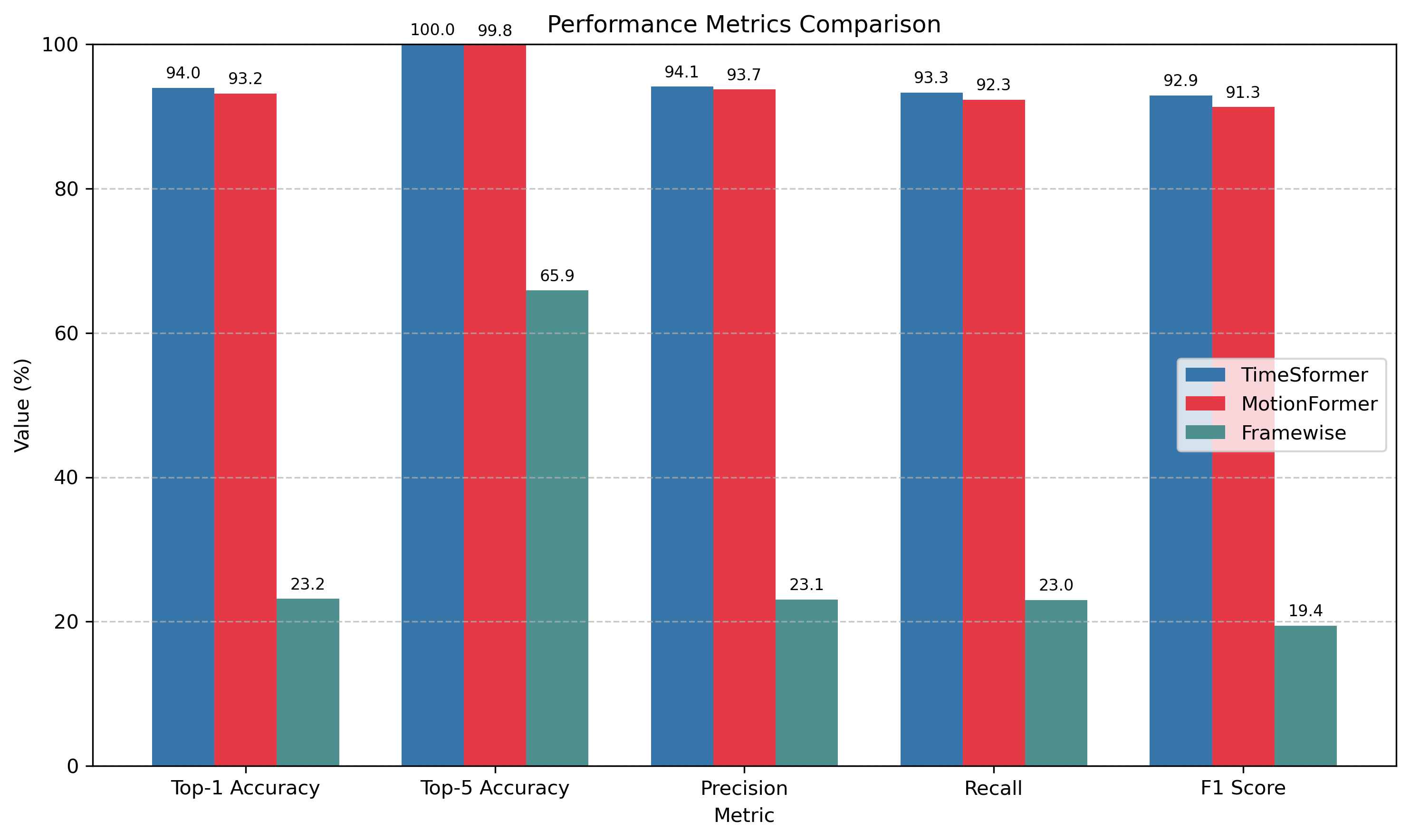}
\caption{Intervention classification performance metrics. Both temporal models substantially outperform the framewise model.}
\label{fig:summary_comparison}
\end{figure}

Summary metrics for all three models (Fig.~\ref{fig:summary_comparison}) demonstrate the stark contrast in performance. Both TSF and MF achieve exceptional top-5 accuracy (100\% for TSF, 99.84\% for MF) and strong macro-averaged precision, recall, and F1 scores above 91\%, indicating robust generalization across the intervention class set.
\section{Future Work and Limitations}
Future work should address several limitations and opportunities:
\begin{enumerate}
    \item \textbf{Clinical representativeness:} The manikin-based approach creates an inherent gap between our simulated environment and NICU settings. Although the interventions were designed to mimic clinical procedures, they lack the variability introduced by infant movement, different body sizes, and unpredictable responses common in real-world care. Future work should validate these models on ethically approved recordings of actual NICU care footage after establishing baseline performance on our controlled Dataset.
    
    \item \textbf{Dataset scope:} The current intervention set of 12 classes excludes clinical procedures and interventions, e.g., respiratory or resuscitation procedures. Expanding the intervention classes would provide a more comprehensive foundation for automated documentation. Furthermore, ICVD treats each intervention as an isolated event, whereas genuine NICU care involves sequences of related interventions. Lastly, it is common for multiple interventions to overlap (e.g., diaper changing and wiping); thus, the problem of intervention detection should likely be framed as a multi-label, rather than multi-class problem.
    
    \item \textbf{Model robustness:} The ICVD recording environment does not fully capture all challenges of actual NICU monitoring systems, which often operate under more variable lighting conditions and more constrained viewing angles than our controlled setting \cite{b12, b3}. Future work should explore model robustness across a broader range of video qualities, including low-resolution streams, and conclusions by clinicians and equipment. Furthermore, since the ICVD contains video from several different cameras and viewpoints, an analysis of classification performance for each camera can inform best practices for actual clinical deployment.
    
    \item \textbf{Clinical translation:} The transition from intervention detection to automated documentation, developing systems that can convert classified interventions into appropriate clinical documentation, including relevant observations and measurements, will require not just accurate classification but also context-aware summarization that aligns with clinical documentation standards and Electronic Health Record (EHR) integration capabilities \cite{b14}.
\end{enumerate}

These directions aim to bridge the gap between technical capability and clinical utility, ultimately working toward systems that can meaningfully reduce the documentation burden.

\section{Conclusion}
We have presented the ICVD, a curated collection of 4,144 videos spanning 12 neonatal care intervention classes, and benchmark performance metrics demonstrating the effectiveness of temporal transformer architectures. The substantial performance gap (70.80\%) between temporal and framewise approaches provides compelling evidence for the necessity of temporal modeling in healthcare intervention classification.

The ICVD addresses a significant data gap in healthcare automation research. With nurses spending 25-50\% of their time on documentation tasks and up to 60\% of routine care interventions going undocumented, the potential impact of automated intervention detection is substantial.

By combining transformer-based video understanding with healthcare domain knowledge, this work establishes a foundation for automated documentation systems that could reduce the administrative burden on NICU staff while improving the completeness and accuracy of patient records. The Dataset's organization mirrors the widely used Kinetics-400 format \cite{b17} and thus facilitates seamless integration with existing deep learning pipelines and supports reproducible research.

The ICVD provides a starting point for bridging the gap between advanced computer vision technologies and clinical documentation needs. As healthcare systems worldwide continue to face staffing challenges and increased documentation requirements, such technological interventions may contribute meaningfully to healthcare efficiency and quality of care. Our work demonstrates the feasibility of using privacy-compliant simulated data to develop effective models that could later be adapted to real clinical environments.

\bibliographystyle{IEEEtran}
\bibliography{bibligraphy}

\end{document}